\documentclass[sigconf]{acmart}

\usepackage{color}
\usepackage{amsmath}
\usepackage{amsfonts}
\usepackage{multirow}

\AtBeginDocument{%
  \providecommand\BibTeX{{%
    \normalfont B\kern-0.5em{\scshape i\kern-0.25em b}\kern-0.8em\TeX}}}

\copyrightyear{2023}
\acmYear{2023}
\setcopyright{acmlicensed}
\acmConference[MMAsia '23]{ACM Multimedia Asia 2023}{December 6--8, 2023}{Tainan, Taiwan}
\acmBooktitle{ACM Multimedia Asia 2023 (MMAsia '23), December 6--8, 2023, Tainan, Taiwan}
\acmPrice{15.00}
\acmDOI{10.1145/3595916.3626399}
\acmISBN{979-8-4007-0205-1/23/12}

\begin{document}
\title{I\textbf{$^2$}SRM: Intra- and Inter-Sample Relationship Modeling for Multimodal Information Extraction}

\author{Yusheng Huang}
\email{huangyusheng@sjtu.edu.cn}
\affiliation{%
  \institution{Shanghai Jiao Tong University}
  \city{Shanghai}
  \country{China}
}

\author{Zhouhan Lin}
\email{lin.zhouhan@gmail.com}
\authornote{Zhouhan Lin is the corresponding author.}
\affiliation{%
  \institution{Shanghai Jiao Tong University}
  \city{Shanghai}
  \country{China}
  }

\renewcommand{\shortauthors}{}

\begin{abstract}
Multimodal information extraction is attracting research attention nowadays, which requires aggregating representations from different modalities. In this paper, we present the Intra- and Inter-Sample Relationship Modeling (\textbf{I}$^2$\textbf{SRM}) method for this task, which contains two modules. Firstly, the intra-sample relationship modeling module operates on a single sample and aims to learn effective representations. Embeddings from textual and visual modalities are shifted to bridge the modality gap caused by distinct pre-trained language and image models. Secondly, the inter-sample relationship modeling module considers relationships among multiple samples and focuses on capturing the interactions. An AttnMixup strategy is proposed, which not only enables collaboration among samples but also augments data to improve generalization. We conduct extensive experiments on the multimodal named entity recognition datasets \emph{Twitter-2015} and \emph{Twitter-2017}, and the multimodal relation extraction dataset \emph{MNRE}. Our proposed method I$^2$SRM achieves competitive results, \textbf{77.12\%} F1-score on \emph{Twitter-2015}, \textbf{88.40\%} F1-score on \emph{Twitter-2017}, and \textbf{84.12\%} F1-score on \emph{MNRE}. \footnote{Codes are available at \href{https://github.com/LUMIA-Group/I2SRM}{https://github.com/LUMIA-Group/I2SRM}}
\end{abstract}



\maketitle

\section{Introduction}\label{intro}
Social media, including Twitter, has become an integral part of our lives, providing a broad platform for individuals or groups to express their understanding and opinions on events \cite{DBLP:conf/naacl/MoonNC18,DBLP:conf/icmcs/ZhengWFF021}. A fundamental and critical step is to unearth hidden insights from social media, which requires information extraction, including named entity recognition \cite{DBLP:journals/tkde/LiSHL22} and relation extraction \cite{DBLP:journals/csur/SmirnovaC19}.

In social media scenarios, information is conveyed not only through texts but also through images. Focusing solely on the text modality may lead to inaccurate information extraction, as validated in \cite{DBLP:journals/corr/abs-2205-03521}. Therefore, Multimodal Information Extraction (MIE) tasks, including Multimodal Named Entity Recognition (MNER) \cite{DBLP:conf/aaai/0001FLH18,DBLP:conf/acl/JiZCLN18} and Multimodal Relation Extraction (MRE) \cite{DBLP:conf/mm/ZhengFFCL021}, have been proposed to effectively obtain structured knowledge from media posts.

\begin{figure}[htb]
  \centering
  \centerline{\includegraphics[width=0.97\linewidth]{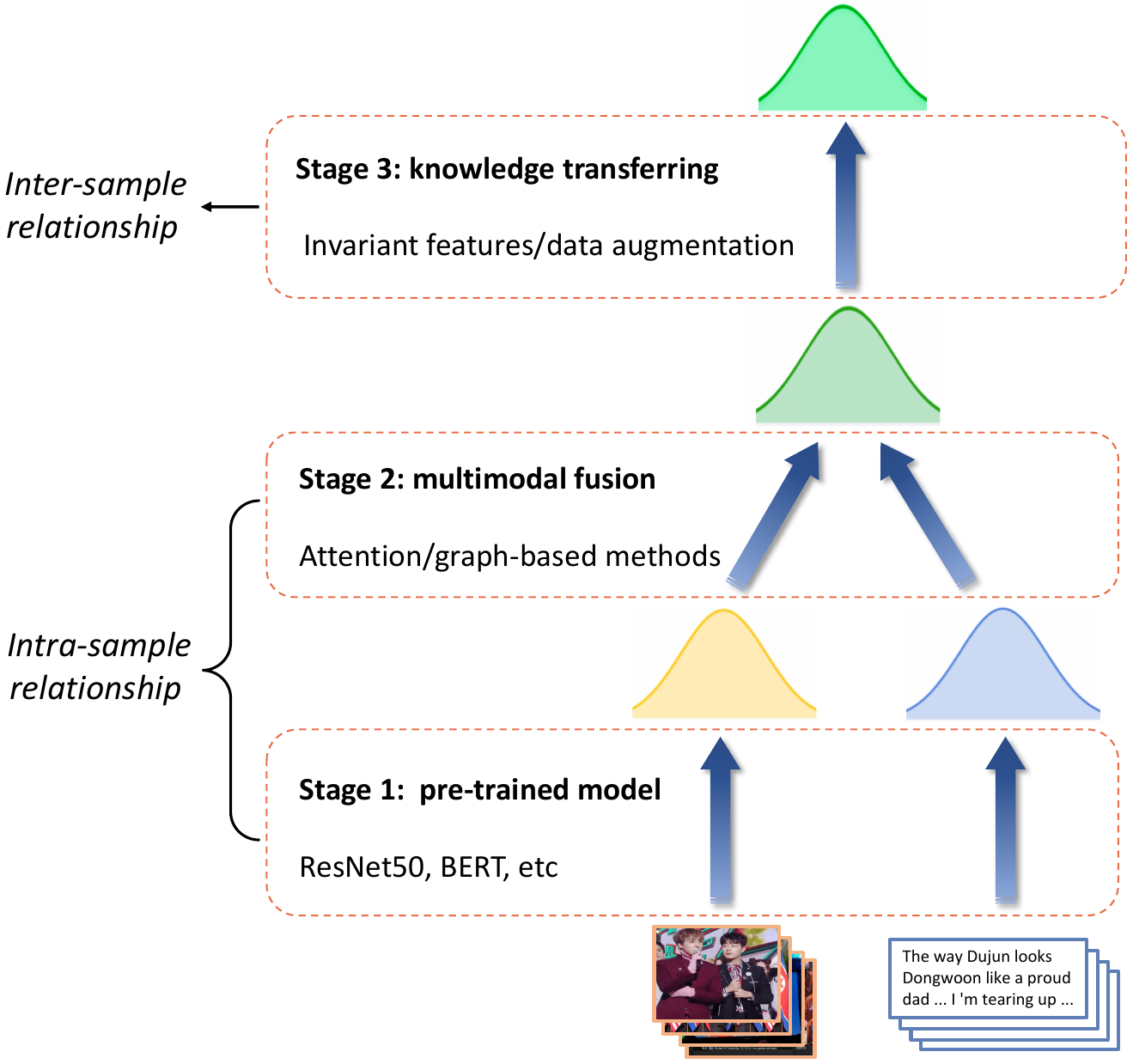}}
  \caption{Formalized workflow for MIE. The intra-sample relationship includes stage 1 and stage 2. The inter-sample relationship includes stag3.}
  \label{fig1_2}
  \Description{flow.}
\end{figure}

As illustrated in Figure \ref{fig1_2}, we first formalize the workflow for MIE, which includes three stages. Specifically,
Stage 1 involves transforming the information from different modalities into embedding vectors, typically initialized with pre-trained models.
Stage 2 focuses on modality fusion, which is elaborately investigated by previous studies.
Stage 3 centers on knowledge transferring, aiming to explore effective methods for representation enhancement, such as invariant feature learning \cite{DBLP:conf/cvpr/HouYT22} and data augmentation \cite{DBLP:conf/iclr/ZhangCDL18}.

In this paper, we focus on stage 1 and stage 3, which are rarely discussed in previous studies.
In stage 1, the primary challenge is the presence of a modality gap between the embeddings of two modalities. Specifically, pre-trained models are trained on separate large-scale datasets, making it difficult to directly capture interactions that are not aligned in the same subspace \cite{DBLP:conf/naacl/WangGJJBWHT22}.
Similar phenomena have been investigated in \cite{liang2022mind,DBLP:conf/emnlp/Ethayarajh19,DBLP:conf/emnlp/LiZHWYL20}, attributing them to different initialization procedures.
In stage 3, the key issue revolves around enhancing the fused representations by learning invariant features from multiple samples or improving model generalization.

To this end, we present a novel Intra- and Inter-Sample Relationship Modeling method dubbed I$^2$SRM for MIE.
I$^2$SRM consists of two modules. 
The first module is the intra-sample relationship modeling module, which operates at the intra-sample level. Its main objective is to bridge the modality gap.  We propose to cast the distribution of object-enhanced textual modality as the conditional prior to regularize the distribution of the visual modality using Kullback–Leibler (KL) divergence. This regularization step improves the alignment between the two modalities, aiding the modality fusion process.
The second module is the inter-sample relationship modeling module, which operates at the multi-sample level. Its main goal is to enhance representations across samples. To achieve this, we propose an AttnMixup strategy that fosters collaboration between samples.
The AttnMixup strategy involves two steps. First, we utilize the multi-head attention mechanism \cite{DBLP:conf/nips/VaswaniSPUJGKP17} to transfer shared knowledge among samples. Then, we enlarge the distribution's vicinal support by leveraging the Mixup \cite{DBLP:conf/iclr/ZhangCDL18} technique to enhance the generalization ability. 

\section{Related Work}\label{related}
\subsection{Multimodal Named Entity Recognition}
With the rapid increase of multimodal information on social media, multimodal named entity recognition has become a fundamental task.
\cite{DBLP:conf/naacl/MoonNC18} introduce a new dataset consisting of short texts and accompanying images and propose a Bi-LSTM-based network with a generic modality-attention module to fuse two modalities.
\cite{DBLP:conf/aaai/0001FLH18} extend the Bi-LSTM network with auxiliary conditional random fields to achieve this task. 
\cite{DBLP:conf/acl/JiZCLN18} introduce a novel visual attention-based model to provide a deeper understanding of images. 
\cite{DBLP:conf/acl/YuJYX20} propose a multimodal interaction module to alleviate the insensitivity issue between the word representation and visual contexts.
\cite{DBLP:conf/dasfaa/ChenLGC21} propose to utilize the attributes and knowledge in images to enhance performance.
\cite{DBLP:conf/mm/WuZCCL020} propose a dense co-attention technique to model the interactions between objects in images and entities in texts.
\cite{DBLP:conf/wsdm/XuHSW22} introduce a general matching and alignment framework to make the representations more consistent.
\cite{DBLP:journals/tmm/ZhengWWCL21} introduce a novel adversarial gated bilinear attention neural network to jointly identify the features from texts and images.
\cite{DBLP:conf/aaai/ZhangWLWZZ21} introduce a unified multi-modal graph fusion method to exploit the semantic correspondences between different semantic units. 
\cite{DBLP:conf/coling/LuZZZ22} introduce a flat multimodal interaction Transformer model to match different modalities.
\cite{DBLP:conf/naacl/WangGJJBWHT22} propose the image-text alignments to align various levels of image features into the textual space.

\subsection{Multimodal Relation Extraction}
Multimodal relation extraction is an important task for obtaining structured information from social media posts.
\cite{DBLP:conf/icmcs/ZhengWFF021} introduce a new multimodal neural relation extraction dataset, which is collected from Twitter. 
\cite{DBLP:conf/mm/ZhengFFCL021} propose a dual graph alignment method to establish the correlation between visual features and textual entities. 
\cite{DBLP:journals/corr/abs-2205-03521} present a hierarchical visual prefix fusion network to enhance the textual representation with the image features. 
\cite{DBLP:journals/ijon/KangLJLZLZ22} introduce a novel translation supervised prototype network for multimodal social relation extraction to extract the features of triples.
\cite{DBLP:conf/coling/0023HDWSSX22} find that different media posts maintain varying degrees of dependence on different modalities. 

While most of the aforementioned approaches primarily concentrate on designing modality fusion mechanisms, our I$^2$SRM method sets itself apart by effectively addressing the modality gap issue and enhancing the fused representations. These essential aspects are often overlooked in MIE.

\begin{figure*}[t]
\centering
\includegraphics[width=0.97\textwidth]{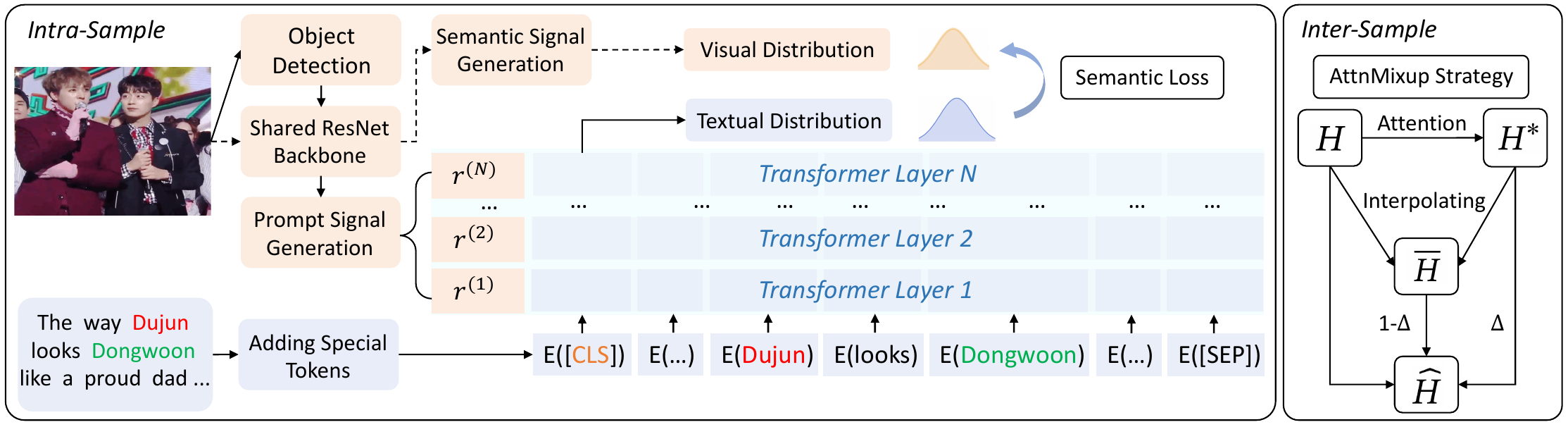}
\caption{The architecture of I$^2$SRM. The approach includes the intra-sample and inter-sample relationship modeling modules. The intra-sample relationship modeling module can rectify different modality distributions. The inter-sample relationship modeling module can enhance representations across samples by AttnMixup Strategy.}
\Description{Architecture.}
\label{framework}
\end{figure*}

\section{Methodology}
An overview is illustrated in Figure \ref{framework}. 
The input to our model consists of a sentence $S$ with $n$ words and a corresponding image $I$. 

\subsection{Intra-Sample Relationship Modeling}\label{intra}
The input sentence $S$ contains a variety of entities $\mathcal{E} = \{e_i\}_{i=1}^p$. 
The input image $I$ contains a series of objects  $\mathcal{O} = \{o_j\}_{j=1}^q$. 

As inspired by \cite{DBLP:conf/mm/ZhengFFCL021,DBLP:conf/wsdm/XuHSW22,DBLP:conf/naacl/WangGJJBWHT22,DBLP:conf/emnlp/ZhaoDSYX022,DBLP:journals/corr/abs-2205-03521} that alignments between objects and entities are critical, we utilize the objects in the image for modality fusion. An advantage is that it can reduce the influence of other noise signals in images.

Therefore, we first detect the image objects by the visual grounding toolkit \cite{DBLP:conf/iccv/YangGWHYL19} and intercept the top $3$ salient objects $\mathcal{O} = \{o_j\}_{j=1}^3$. These objects are further resized to $224 \times 224$ pixels.
Then, we feed these detected objects to the Pre-trained Image Model (PIM) such as the ResNet50 \cite{DBLP:conf/cvpr/HeZRS16} backbone, 
\begin{equation}\label{eq_pim1}
    \boldsymbol{p_j} = \text{PIM} (o_j), j = 1, 2, 3,
\end{equation}
where $\boldsymbol{p_j}$ denotes the corresponding object embedding.

We employ the Pre-trained Language Model (PLM) such as BERT \cite{DBLP:conf/naacl/DevlinCLT19} and RoBERTa \cite{DBLP:journals/corr/abs-1907-11692} to generate the contextualized representation of the input sentence $S$,
\begin{equation}
    \boldsymbol{e_s^{(l)}} = \text{PLM} (S), l = 1, 2, ..., N,
\end{equation}
where $\boldsymbol{e_s^{(l)}}$ denotes the sentence embedding in the $l_{th}$ Transformer layer with $N$ layers in total.

For the fusion method of multimodal features, the multi-level fusion technique is an effective and widely used method as multiple levels of features can be captured \cite{DBLP:conf/cvpr/YeR0W19}.
Typically, given the object embedding $\boldsymbol{p_j}$, the multi-level features $\boldsymbol{r_j^{(l)}}, l = 1, 2, ..., N$ are generated by extracting the hidden states of different backbone blocks.
An average pooling is adopted to adjust the spatial size,
\begin{equation}
    \boldsymbol{r^{{(l)}}} = \text{Conv$_{1\times 1}$} ( \text{avg-pool$_j$} ( \boldsymbol{p_j^{(l)}})), l = 1, 2, ..., N,
\end{equation}
where the 1$\times$1 convolutional layer aims to make the channel dimensions of various levels consistent.

We call $\boldsymbol{r^{{(l)}}}, l = 1, 2, ..., N$ {prompt signal} following \cite{DBLP:journals/corr/abs-2110-07602}, and they provide disambiguation information and visual signals for the entities and the textual modality in a hierarchical structure.
Then, as shown in Figure \ref{framework}, we prepend the prompt signal to the sentence embedding in PLM Transformer layers to aggregate multimodal features,
\begin{equation}\label{eq_gap}
    \boldsymbol{M_N} = \text{Transformers} ( \text{Concat} [\boldsymbol{r^{{(l)}}}; \boldsymbol{e_s^{(l)}}]),
\end{equation}
where $\boldsymbol{M_N}$ denotes the hidden states of the $N_{th}$ Transformer layer.

The image embedding $\boldsymbol{e_I}$ is generated by feeding the image $I$ to the pre-trained image model,
\begin{equation}\label{eq_pim2}
    \boldsymbol{e_I} = \text{PIM} (I).
\end{equation}

Notably, Eq. \ref{eq_pim1} and Eq. \ref{eq_pim2} share the same backbone.

We call $\boldsymbol{e_I}$ {semantic signal} because it provides the overall semantic information of visual modality.
Particularly, the modality gap lies in the different distributions of different modalities, which is introduced in Eq. \ref{eq_gap}.
This is inevitable as different pre-trained models are utilized \cite{liang2022mind}.

The core idea to alleviate the modality gap issue is to cast the distribution of the primary object-enhanced textual modality as a conditional prior, and then regularize the distribution of the visual modality \cite{DBLP:conf/coling/0023HDWSSX22,DBLP:conf/iclr/JoySTRSN22}.
We employ $\boldsymbol{M_N}$ as a prior, which is generated from the combination of the object embedding and the text embedding.
Concretely, the distribution of $\boldsymbol{M_N}$ is denoted as $p(\gamma)$,
\begin{equation}\label{eq_recM1}
    p(\gamma) \sim f_{\psi} ( \boldsymbol{M_N} ),
\end{equation}
where $\psi$ denotes the parameters of a linear projection $f$.

The semantic loss $\mathcal{L}_{sem}$ is as calculated based on the KL-divergence,
\begin{equation}\label{eq_recM2}
    \mathcal{L}_{sem} = \text{KL}( p(\gamma) || p(\beta) ). 
\end{equation}

Notably, during the calculation of the {semantic signal} and {prompt signal}, a shared visual backbone is employed. Visual features of different levels have been utilized.

\subsection{Inter-Sample Relationship Modeling}\label{inter}
Given the mini-batch sample embedding, we propose an AttnMixup strategy to explore the inter-sample relationships, which could further enhance the representations. 

Firstly, the training set of MIE can be imperfect, and the number of samples in different categories varies greatly \cite{DBLP:conf/mm/ZhengFFCL021}.
To this end, we leverage the multi-head attention mechanism \cite{DBLP:conf/nips/VaswaniSPUJGKP17} to transfer knowledge among sample embeddings \cite{DBLP:conf/cvpr/HouYT22}.
Specifically, for the mini-batch sample embedding $H \in \mathbb{R}^{B\times C} = \{ \boldsymbol{M_N}^{(i)} \in \mathbb{R^C}, i = 1, 2, ..., B\}$, we transform $H$ as follows, 
\begin{equation}
    r_{ij} = \mathbf{v}^T \tanh ( \mathbf{W} \cdot H + \mathbf{b} ),
\end{equation}
\begin{equation}
    \alpha_{ij} = {\rm softmax}(r_{ij}) = \frac{\exp \left(r_{ij}\right)}{\sum_{k=1}^{C} \exp \left(r_{ik}\right)},
\end{equation}
\begin{equation}
    H^{\prime}= \sum_{j=1}^{C} { \alpha }_{ij} * H,
\end{equation}
where $\mathbf{W}$, $\mathbf{v}$, and $\mathbf{b}$ are parameters to be learned. 

Next, we adopt the multi-head mechanism to capture features from different aspects, where invariant features among samples can be learned,
\begin{equation}\label{eq_Attnhead}
    \text{Multi-head}(H) = \text{Concat}( H_1^{\prime};   H_2^{\prime}; ...;  H_k^{\prime}),
\end{equation}
where $k$ is the number of heads.
Then, we adopt the residual connection and layer normalization to obtain the normalized mini-batch embedding $H^* \in \mathbb{R}^{B\times C}$ as follows, 
\begin{equation}\label{eq_ln}
    H^* = \text{LayerNorm}(\text{Multi-head}(H)\cdot \mathbf{W^d} +H),
\end{equation}
where $\mathbf{W^d}$ denotes the weight to align the dimensionality.

Notably, the $H$ and transformed $H^*$ possess the same target labels, which can be used together for training. 

Secondly, we augment $H$ and $H^*$ by the vicinal risk minimization principle \cite{DBLP:conf/nips/ChapelleWBV00,DBLP:conf/iclr/ZhangCDL18,DBLP:conf/aaai/Guo20,DBLP:journals/corr/abs-1905-08941}. 
We specify a new sampling set $\widetilde{H}$ constructed from $H$ and $H^*$,
\begin{equation}\label{eq_delta}
    \widetilde{H} = \delta \cdot H  \cup (1-\delta ) \cdot H^* ,
\end{equation}
where $\delta \in [0, 1]$ refers to the sampling ratio.

Then, we generate the synthetic sample set $\overline{H}$ by linearly interpolating a pair of sample embeddings and the corresponding targets in $\widetilde{H}$.
Concretely, given a pair of samples $(x_i, y_i)$ and $(x_j, y_j)$ in $\widetilde{H}$, where $x_i$ and $x_j$ denote the input embedding and $y_i$ and $y_j$ denote their corresponding target labels, we select a synthesis ratio $\lambda$ from the Beta distribution, 
and the synthetic sample $(x_k, y_k)$ are generated as follows,
\begin{equation}
    x_k = \lambda \cdot x_i + (1 - \lambda) \cdot x_j ,
\end{equation}
\begin{equation}
    y_k = \lambda \cdot y_i + (1 - \lambda) \cdot y_j .
\end{equation}

The final training set $\widehat{H}$ includes three parts: the original samples in $H$, the transformed samples in $H^*$ to transfer knowledge and learn invariant features, and the synthetic samples in $\overline{H}$ to enlarge the vicinal support of the training distribution,
\begin{equation}\label{eq_Delta}
    \widehat{H} = H \cup \Delta \cdot H^*  \cup (1-\Delta ) \cdot \overline{H} ,
\end{equation}
where $\Delta \in [0, 1]$ aims to adjust the ratio.

The embedding of test samples can be directly used for prediction without the above process, i.e., without generating transformed samples and synthetic samples. In this way, we can avoid the requirement of layer normalization on batch statistics in Eq. \ref{eq_ln}.

\subsection{Classifier}\label{classifier}
\noindent\textbf{Multimodal Named Entity Recognition.}
For the named entity recognition task, it is critical to consider the constraints of target labels in the neighbors.
Studies have shown that Conditional Random Fields (CRFs) \cite{DBLP:conf/icml/LaffertyMP01} maintain the ability to explore the correlations for sequence labeling \cite{DBLP:conf/aaai/0001FLH18,DBLP:conf/acl/YuJYX20}.
We follow the implementation of CRFs in \cite{DBLP:conf/acl/MaH16} and the input embeddings in $\widehat{H}$ are fed to CRFs.

For a sequence of target labels $y= \{y^{(1)}, y^{(2)}, ..., y^{(n)} \}$, we jointly minimize the summation of the semantic loss $\mathcal{L}_{sem}$ and the negative log-likelihood of the target labels $y$ as follows, 
\begin{equation}\label{eq_sem1}
 \mathcal{L}_{\mathrm{NER}} = -\frac{1}{n} \sum_{i=1}^n \log p(y^{(i)}| \widehat{H} ) + \lambda_{sem} \cdot \mathcal{L}_{sem},
\end{equation}
where $y$ follows the pre-defined BIO tagging schema. $\lambda_{sem}$ denotes the parameter to adjust the ratio between two losses.

\noindent\textbf{Multimodal Relation Extraction.}
The multimodal relation extraction task is formulated as a classification task, which aims to predict the relation types between the subject and the object entities.
We use the concatenation of head and tail entity embeddings as the sample embedding,
and we leverage the softmax function to compute the probability distribution,
\begin{equation}
    e_{pair} = \text{softmax} ( \text{Concat} ( \widehat{H}_{[head]} ; \widehat{H}_{[tail]} ) ) .
\end{equation}

The final loss is the summation of the cross-entropy loss and the semantic loss $\mathcal{L}_{sem}$, 
\begin{equation}\label{eq_sem2}
 \mathcal{L}_{\mathrm{RE}} = - \log p(r| e_{pair} ) + \lambda_{sem} \cdot \mathcal{L}_{sem},
\end{equation}
where $r$ denotes the pre-defined relation types between entities. $\lambda_{sem}$ is a parameter used to adjust the ratio between two losses.

\begin{table*}
\caption{Comparison of performance ($\%$) with state-of-the-art models. The best results are in bold. }
\centering
\resizebox{0.95\textwidth}{!}{
\begin{tabular}{c   l   ccc   ccc   ccc}
\toprule
\multirow{2}{*}{\textbf{Modality}}  &    \multirow{2}{*}{\textbf{Model}}  & \multicolumn{3}{c}{\textbf{Twitter-2015}} &   \multicolumn{3}{c}{\textbf{Twitter-2017}}  &   \multicolumn{3}{c}{\textbf{MNRE}} \\
&       & Precision & Recall & F1 & Precision & Recall & F1 & Precision & Recall & F1 \\
\midrule
\multirow{6}{*}{Text}    &   {CNN-BiLSTM-CRF} \cite{DBLP:conf/acl/MaH16}  & 66.24 & 68.09 & 67.15 & 80.00 & 78.76 & 79.37 & - & - & - \\
&   {BiLSTM-CRF} \cite{DBLP:conf/naacl/LampleBSKD16} & 70.32 &  68.05 &  69.17 &  82.69  & 78.16  & 80.37 & - & -  & - \\
&   {BERT} \cite{DBLP:conf/naacl/DevlinCLT19} & 68.30  & 74.61 &  71.32 &  82.19 &  83.72  & 82.95  & - & -  & - \\
&   {BERT-CRF} & 69.22  & 74.59 &  71.81 &  83.32 &  83.57  & 83.44  & - & -  & - \\
&   {PCNN}  \cite{DBLP:conf/emnlp/ZengLC015} & - & -  & - &  - & -  & -  & 62.85   & 49.69 &  55.49 \\
&   {MTB} \cite{DBLP:conf/acl/SoaresFLK19} & - & -  & - &  - & -  & -  & 64.46  & 57.81 &  60.86 \\
\midrule
\multirow{11}{*}{Text+Image}    &   VisualBERT \cite{DBLP:journals/corr/abs-1908-03557} & 68.84 & 71.39 & 70.09 & 84.06 & 85.39 & 84.72 & 57.15 & 59.48 & 58.30 \\
&   AdapCoAtt-BERT-CRF \cite{DBLP:conf/aaai/0001FLH18} & 69.87 &  74.59 &  72.15 &  85.13  & 83.20  & 84.10 & - & -  & - \\
&   UMT \cite{DBLP:conf/acl/YuJYX20} & 71.67  & 75.23 &  73.41 &  85.28 &  85.34  & 85.31  & 62.93   & 63.88 &  63.46 \\
&   MEGA \cite{DBLP:conf/mm/ZhengFFCL021} & 70.35 & 74.58  & 72.35 & 84.03 & 84.75 &  84.39 & 64.51 & 68.44 &  66.41  \\
&   UMGF \cite{DBLP:conf/aaai/ZhangWLWZZ21} & 74.49 &  75.21 &  74.85 &  86.54 &  84.50 &  85.51  & 64.38  & 66.23  & 65.29  \\
&   MAF \cite{DBLP:conf/wsdm/XuHSW22} & 71.86 & 75.10 & 73.42 & 86.13 & 86.38 & 86.25 & - & - & - \\
&   HVPNeT \cite{DBLP:journals/corr/abs-2205-03521} & 73.87 & 76.82 & 75.32 & 85.84 & 87.93 & 86.87 & 83.64 & 80.78 & 81.85 \\
&   ITA \cite{DBLP:conf/naacl/WangGJJBWHT22} & - & - & 76.01 & - & - & 86.45 & - & -  & - \\
&   \textbf{I$^2$SRM} & \textbf{76.22} & \textbf{78.04} & \textbf{77.12} & \textbf{87.41} & \textbf{89.42} & \textbf{88.40} & \textbf{84.65} & \textbf{83.59} & \textbf{84.12} \\
\bottomrule
\end{tabular}}
\label{result1}
\end{table*}

\section{Experiments} 
\subsection{Datasets and Metrics}
We conduct experiments on two multimodal information extraction tasks, namely multimodal named entity recognition (MNER) and multimodal relation extraction (MRE).
For the MNER task, we adopt two widely used datasets, Twitter-2015 \cite{DBLP:conf/aaai/0001FLH18} and Twitter-2017 \cite{DBLP:conf/acl/JiZCLN18}, which are collected from Twitter.
Particularly, we use the pre-processed datasets provided by \cite{DBLP:journals/corr/abs-2205-03521}, which modified some mis-annotations in Twitter-2015. 
The Twitter-2015 dataset contains $4,000$ samples for training, $1,000$ samples for validation, and $3,257$ samples for testing.
The Twitter-2017 dataset contains $4,290$ samples for training, $1,432$ samples for validation, and $1,459$ samples for testing.
For the MRE task, we adopt the MNRE \cite{DBLP:conf/mm/ZhengFFCL021} dataset, which is constructed from user tweets. 
The MNRE dataset contains $23$ pre-defined relation types. There are $9,201$ samples in total and $12,247$ entity pairs for training, $1,624$ entity pairs for validation, and $1,614$ entity pairs for testing, respectively.

Following previous studies \cite{DBLP:journals/corr/abs-2205-03521,DBLP:conf/acl/JiZCLN18,DBLP:conf/mm/ZhengFFCL021}, we use the micro precision, recall, and F1-score to evaluate the model performance.

\subsection{Parameter Settings}
During training, we leverage PyTorch $1.7.1$ to conduct experiments with NVIDIA A100 GPUs.  
We finetune our model using the AdamW \cite{DBLP:conf/iclr/LoshchilovH19} optimizer with a linear warmup for the first $6\%$ steps followed by a linear decay to $0$. 
The batch size is $8$ for MNER and $16$ for MRE. 
The number of attention heads is $4$ in Eq. \ref{eq_Attnhead} followed by a dropout layer with the rate being $0.2$.
We set $\delta=0.7$ in Eq. \ref{eq_delta} to specify the sampling set.
We set $\Delta=0.6$ in Eq. \ref{eq_Delta} to aggregate the final training set.
$\lambda_{sem}$ is set to be $0.5$ in Eq. \ref{eq_sem1} and Eq. \ref{eq_sem2}.
Our models are trained for $30$ epochs and $25$ epochs for MNER and MRE, respectively. 
All of the reported results are averaged from $3$ runs using different random seeds. 
In both MNER and MRE tasks, we select the best-performing models on the validation sets and evaluate them on the test sets.

\subsection{Baselines}
We compare our I$^2$SRM model with the following baselines for a comprehensive comparison.
The baselines consist of two groups.

The first group contains these text-based approaches:
\begin{itemize}
    \item \textbf{CNN-BiLSTM-CRF} \cite{DBLP:conf/acl/MaH16} : It uses the combination of LSTM, CNN, and CRF to achieve the sequence labeling task.
    \item \textbf{HBiLSTM-CRF} \cite{DBLP:conf/naacl/LampleBSKD16} : It models the output label dependencies to explicitly build and label chunks of the inputs.
    \item \textbf{BERT} \cite{DBLP:conf/naacl/DevlinCLT19} : It is finetuned using the pre-trained BERT model with an additional softmax layer for prediction.
    \item \textbf{BERT-CRF} : It leverages the pre-trained BERT model with the CRF output layer.
    \item  \textbf{PCNN}  \cite{DBLP:conf/emnlp/ZengLC015} : It designs a piecewise convolutional neural network to automatically learn relevant features.
     \item \textbf{MTB} \cite{DBLP:conf/acl/SoaresFLK19} : It builds the task agnostic relational representations from entity-linked texts based on BERT.
\end{itemize}

The second group contains these competitive multimodal approaches: 
\begin{itemize}
    \item \textbf{VisualBERT} \cite{DBLP:journals/corr/abs-1908-03557} : It uses Transformer layers to implicitly align elements of texts and regions in images.
    \item \textbf{AdapCoAtt-BERT-CRF}  \cite{DBLP:conf/aaai/0001FLH18} : It proposes an adaptive co-attention network for modality fusion.
    \item \textbf{UMT} \cite{DBLP:conf/acl/YuJYX20} : It introduces a unified multimodal Transformer to guide the extractions with entity span predictions.
    \item \textbf{MEGA} \cite{DBLP:conf/mm/ZhengFFCL021} : It employs a dual graph to capture the correlations between visual objects and textual expressions.
    \item  \textbf{UMGF} \cite{DBLP:conf/aaai/ZhangWLWZZ21} : It uses a unified multimodal graph fusion approach to capture the semantic correspondences.
    \item \textbf{MAF} \cite{DBLP:conf/wsdm/XuHSW22} : It introduces a general matching and alignment method to bridge the gap between two modalities.
    \item \textbf{HVPNeT} \cite{DBLP:journals/corr/abs-2205-03521} : It builds a novel hierarchical visual prefix fusion network to enhance the fused representations.
    \item \textbf{ITA} \cite{DBLP:conf/naacl/WangGJJBWHT22} : It aligns the image features into the textual spaces to better exploit the Transformer-based embeddings.
\end{itemize}

\begin{table}
\centering
\caption{F1-score (\%) on Twitter-2015 test set with different pre-trained models. The best result is in bold.}
\resizebox{0.83\columnwidth}{!}{
\begin{tabular}{lcccc}
\toprule
\textbf{Model}  & ResNet50 & ResNet101 & ResNet152 \\
\midrule
BERT-base    &  76.65  & 76.50  &  75.42  \\
RoBERTa-base  & 76.70  & 76.51   &  75.70  \\
RoBERTa-large   & \textbf{77.12}  & 76.89 &  76.50  \\
\bottomrule
\end{tabular}}
\label{twitter15}
\end{table}

\begin{table}
\centering
\caption{F1-score (\%) on MNRE Twitter-2017 set using different pre-trained models. The best result is presented in bold.}
\resizebox{0.83\columnwidth}{!}{
\begin{tabular}{lcccc}
\toprule
\textbf{Model}  & ResNet50 & ResNet101 & ResNet152 \\
\midrule
BERT-base    & 88.16   & 87.70   &  87.41   \\
RoBERTa-base  &  88.29   &  88.02  &   87.53  \\
RoBERTa-large   & \textbf{88.40}  &  88.32  &  87.89   \\
\bottomrule
\end{tabular}}
\label{twitter17}
\end{table}

\begin{table}
\centering
\caption{F1-score (\%) on MNRE test set utilizing different pre-trained models. The best result is shown in bold.}
\resizebox{0.83\columnwidth}{!}{
\begin{tabular}{lcccc}
\toprule
\textbf{Model}  & ResNet50 & ResNet101 & ResNet152 \\
\midrule
BERT-base    & 83.78  & 83.41 & 83.17  \\
RoBERTa-base  & 84.08  & 83.83 & 83.51  \\
RoBERTa-large   & \textbf{84.12}  & 83.90 & 83.69   \\
\bottomrule
\end{tabular}}
\label{mnre}
\end{table}

\subsection{Effect of Pre-trained Models}\label{effectPLM}
We first clarify why the pre-trained multimodal models such as VisualBERT \cite{DBLP:journals/corr/abs-1908-03557}, ViLBERT \cite{DBLP:conf/nips/LuBPL19} and Unicoder-VL \cite{DBLP:conf/aaai/LiDFGJ20}  are not used directly, as there is a modal gap issue with different pre-trained models trained on different datasets.
The reason is that for MIE tasks, social media posts are not making predictions on the image side \cite{DBLP:journals/corr/abs-2205-03521}.
This is exactly what distinguishes the MIE tasks from other multimodal tasks such as visual question answering.
Therefore, using pre-trained multimodal models could be suboptimal.
For MIE, images might not be as significant as texts \cite{DBLP:conf/coling/0023HDWSSX22}, and images are usually used to enhance the overall representation \cite{DBLP:conf/wsdm/XuHSW22,DBLP:conf/mm/ZhengFFCL021}.

We then investigate one question, i.e., should we work harder on PLM or PIM to improve the performance of MIE?
To this end, we select three pre-trained models with rather similar structures for PLM and PIM respectively, namely BERT-base \cite{DBLP:conf/naacl/DevlinCLT19}, RoBERTa-base \cite{DBLP:journals/corr/abs-1907-11692}, RoBERTa-large \cite{DBLP:journals/corr/abs-1907-11692} for PLM (sharing the same model structure), and ResNet50 \cite{DBLP:conf/cvpr/HeZRS16}, ResNet101 \cite{DBLP:conf/cvpr/HeZRS16}, ResNet152 \cite{DBLP:conf/cvpr/HeZRS16} for PIM (sharing the same model block).
We conduct experiments using the aforementioned backbones on Twitter-2015, Twitter-2017, and MNRE. The experimental results are shown in Table \ref{twitter15}, Table \ref{twitter17}, and Table \ref{mnre}, respectively.
We can observe that leveraging larger PLM is beneficial since it maintains the ability to consistently improve performance.
However, the performance will be reduced when deeper PIM is used. 
The reason might be that deeper PIMs are more difficult to finetune with PLMs simultaneously, and larger parameter quantities make it difficult to shift the distribution. Besides, results indicate that the feature extraction capability of ResNet50 is sufficient.
Therefore, we employ RoBERTa-large as the text encoder and leverage ResNet50 as the image backbone in the following experiments based on these discussions.
Overall, results demonstrate that using superior PLMs is beneficial for MIE tasks.

\subsection{Main Results}
Table \ref{result1} presents the performance comparison with various text-based approaches and competitive multimodal approaches.
From the experimental results, we can observe that I$^2$SRM achieves the highest F1-score on three datasets, $77.12\%$ F1-score on Twitter-2015, $88.40\%$ F1-score on Twitter-2017, and $84.12\%$ F1-score on MNRE.
These results verify the effectiveness of bridging the modality gap and enhancing representations using the AttnMixup Strategy.

Compared with the text-based approaches, Our I$^2$SRM model maintains a significant advantage. Compared with BERT-CRF, our I$^2$SRM improves the F1-score by $5.31\%$ on Twitter-2015 and improves the F1-score by $4.96\%$ on Twitter-2017. Compared with MTB, our I$^2$SRM improves the F1-score by $23.26\%$ on MNRE. These results demonstrate visual information is beneficial for MIE tasks, especially the MRE task.

Compared with the multimodal approaches, our I$^2$SRM model achieves the best performance on all evaluation criteria across three datasets.
Compared with ITA, our model does not require explicit alignment of features in multiple dimensions such as regional object tags, image-level captions, and optical characters. 
Compared with VisualBERT, our model has improved by $7.03\%$, $3.68\%$, and $25.82\%$ F1-score on Twitter-2015, Twitter-2017, and MNRE, respectively.
The results suggest that the current multimodal pre-trained model is not suitable for MIE, as we have illustrated in \S \ref{effectPLM}.
Compared with other approaches, our model verifies the effectiveness of leveraging intra- and inter-sample relationships.

\begin{table}
\centering
\caption{F1-score (\%) on the test set for ablation study. }
\begin{tabular}{lccc}
\toprule
\textbf{Model}  & Twitter-2015 & Twitter-2017 & MNRE \\
\midrule
I$^2$SRM   & \textbf{77.12}  & \textbf{88.40} & \textbf{84.12} \\
\midrule 
- Semantic Loss $\mathcal{L}_{sem}$  & 76.43 & 87.85 & 83.42 \\
- AttnMixup Strategy & 76.57 & 87.92 & 83.36 \\
\bottomrule
\end{tabular}
\label{ablation}
\end{table}

\subsection{Ablation Study}
To further investigate the effectiveness of our model, we perform a comparison between I$^2$SRM and its ablations concerning two critical components, i.e., the semantic loss $\mathcal{L}_{sem}$ and the AttnMixup strategy. 

Table \ref{ablation} presents the results for the ablation study on three datasets. 
For the Twitter-2015 MNER dataset, removing the semantic loss $\mathcal{L}_{sem}$ will result in a $0.69\%$ F1-score decrease, while removing the AttnMixup strategy will result in a $0.55\%$ F1-score decrease.
This indicates that rectifying the modality gap is more important for this dataset.
For the Twitter-2017 MNER dataset, removing the semantic loss $\mathcal{L}_{sem}$ leads to a $0.55\%$ performance drop, and removing the AttnMixup strategy leads to a $0.48\%$ performance drop. These results follow the same trend as on Twitter-2015.
We believe that the modality gap issue has a greater impact on model performance for the MNER task.
For the MNRE dataset, removing the semantic loss $\mathcal{L}_{sem}$ will result in a $0.70\%$  F1-score decrease, and removing the AttnMixup strategy will result in a $0.76\%$  F1-score decrease.
Different from the MNER task, the AttnMixup strategy has a greater impact on the model performance. We speculate that the reason is that the size of the MNRE dataset is not large. Therefore, the data augmentation has a greater impact on performance.

Overall, results show that both components contribute to the final improvements on Twitter-2015, Twitter-2017, and MNRE datasets.

\begin{table}
\centering
\caption{F1-score (\%) on the test set for parameter sensitivity study with I$^2$SRM. }
\begin{tabular}{lccc}
\toprule
\textbf{$\Delta$}  & Twitter-2015 & Twitter-2017 & MNRE \\
\midrule
0.8   & 76.53  &  87.92 &  83.62 \\
0.6   & \textbf{77.12}  & \textbf{88.40} & \textbf{84.12} \\
0.4   & 75.95  &  87.71 & 81.68  \\
\bottomrule
\end{tabular}
\label{sensitivity}
\end{table}

\subsection{Parameter Sensitivity Study}
We evaluate our model using different parameter settings. 
We focus on one key parameter $\Delta$ in Eq. \ref{eq_Delta}. In the inter-sample relationship modeling module, the parameter $\Delta$ controls the proportion of samples generated in different ways in the final training set, i.e., $H^*$ generated using the multi-head attention mechanism and $\overline{H}$ generated by synthesizing samples.

Table \ref{sensitivity} presents the results of our I$^2$SRM model influenced by different values of $\Delta$.
We can observe that when $\Delta$ equals $0.8$ and $0.6$, our I$^2$SRM model can achieve comparable results across three datasets. 
However, when $\Delta$ equals $0.4$, the model performance declines to a lower level on three datasets, especially on the MNRE dataset, achieving $81.68\%$ F1-score on the test set. 
We speculate that this might be due to the following reasons.
Firstly, samples generated through the multi-head attention mechanism could better preserve the properties of the original samples.
The generated samples can learn invariant features from the original samples, which could enhance the representations.
Secondly, samples generated by linear interpolation can enlarge the vicinal support of the training distribution. 
However, noise labels are introduced in this process.
Therefore, if the number of noisy samples exceeds a certain percentage, the model optimization would be greatly troubled by these noise samples.
The reason for the greater impact on the MNRE dataset might be the smaller scale of entity pairs.

\section{Conclusion} 
In this paper, we investigate the intra- and inter-sample relationships for multimodal information extraction. We formalize the workflow, highlighting the modality gap issue and the absence of a knowledge transferring tier.
To address these challenges, we propose a novel I$^2$SRM model, consisting of two modules. 
The first module, the intra-sample relationship modeling module, operates at the single-sample level and effectively bridges the modality gap using a semantic loss based on the Kullback–Leibler divergence. 
The distribution of object-enhanced textual modality is utilized as a conditional prior to regularize the distribution of the visual modality.
The second module, the inter-sample relationship modeling module, operates at the multi-sample level and capitalizes on our designed AttnMixup strategy to enhance the fused representations and improve model generalization.
Notably, both the semantic loss and AttnMixup strategy can serve as plug-and-play parts for other models.
Extensive experimental results have verified the effectiveness of our I$^2$SRM model, showcasing its capability to improve the performance of multimodal information extraction tasks.

\begin{acks}
The  authors  would  like  to  thank  the  support  from the National Natural Science Foundation of China (NSFC) grant (No. 62106143), and Shanghai Pujiang Program (No. 21PJ1405700).
\end{acks}

\bibliographystyle{ACM-Reference-Format}
\bibliography{sample-base}

\end{document}